\documentclass[twocolumn, switch]{article}
\usepackage{arxiv}

\usepackage[utf8]{inputenc}

\usepackage{cite}
\usepackage{amsmath,amssymb,amsfonts}
\usepackage{algorithmic}
\usepackage{graphicx}
\usepackage{textcomp}
\usepackage{xcolor}
\usepackage{subcaption}
\usepackage{multirow}
\usepackage{booktabs,threeparttable}
\usepackage{longtable}
\usepackage{array}
\usepackage{ragged2e}
\usepackage{arydshln}
\usepackage{authblk}

\usepackage{utf8}
\novocalize
\setfarsi
\settrans{farsi}
\yahnodots

\newsavebox{\spacebox}
\begin{lrbox}{\spacebox}
\verb*! !
\end{lrbox}
\newcolumntype{L}{>{\raggedright\arraybackslash}X}
\newcommand{\filliation}[5]{\affil[#1]{\textbf{#2}\vskip 0pt \textbf{#3}\vskip 0pt #4\vskip 0pt #5\vspace{10pt}}} 

\title{Leveraging ParsBERT and Pretrained mT5 for Persian Abstractive Text Summarization}

\author[1]{Mehrdad Farahani}
\author[2]{Mohammad Gharachorloo}
\author[3]{Mohammad Manthouri}

\filliation{1}{Dept. of Computer Engineering}{Islamic Azad University North Tehran Branch}{Tehran, Iran}{m.farahani@iau-tnb.ac.ir}
\filliation{2}{Dept. of Electrical Engineering and Robotics}{Queensland University of Technology}{Brisbane, Australia}{mohammad.gharachorloo@connect.qut.edu.au}
\filliation{3}{Dept. of Electrical and Electronic Engineering}{Shahed Univerisity}{Tehran, Iran}{mmanthouri@shahed.ac.ir}

\begin{document}

\twocolumn[ 
  \begin{@twocolumnfalse} 
  
\maketitle

\begin{abstract}
Text summarization is one of the most critical Natural Language Processing (NLP) tasks. More and more researches are conducted in this field every day. Pre-trained transformer-based encoder-decoder models have begun to gain popularity for these tasks. This paper proposes two methods to address this task and introduces a novel dataset named pn-summary for Persian abstractive text summarization. The models employed in this paper are mT5 and an encoder-decoder version of the ParsBERT model (i.e., a monolingual BERT model for Persian). These models are fine-tuned on the pn-summary dataset. The current work is the first of its kind and, by achieving promising results, can serve as a baseline for any future work.
\end{abstract}
\keywords{Text Summarization \and Abstractive Summarization \and Pre-trained Based \and BERT \and mT5}
\vspace{0.35cm}
\end{@twocolumnfalse}
]

\section{Introduction}
With the emergence of the digital age, a vast amount of textual information has become digitally available. Different Natural Language Processing (NLP) tasks focus on different aspects of this information. Automatic text summarization is one of these tasks and concerns about compressing texts into shorter formats such that the most important information of the content is preserved \cite{nenkova2012survey, edmundson1969new}. This is crucial in many applications since generating summaries by humans, however precise, can become quite a time consuming and cumbersome. Such applications include text retrieval systems used in search engines to display a summarized version of the search results \cite{turpin2007fast}.\\
Text summarization can be viewed from different perspectives including single-document \cite{patil2015automatic} vs. multi document\cite{christensen2013towards, nenkova2006compositional} and monolingual vs. multi-lingual \cite{gambhir2017recent}. However, an important aspect of this task is the approach, which is either extractive or abstractive. In extractive summarization, a few sentences are selected from the context to represent the whole text. These sentences are selected based on their scores (or ranks). These scores are determined by computing certain features such as the ordinal position of sentences concerning one another, length of the sentence, a ratio of nouns, etc. After sentences are ranked, the top n sentences are selected to represent the whole text \cite{gupta2010survey}.\\
Abstractive summarization techniques create a short version of the original text by generating new sentences with words that are not necessarily found in the original text. Compared to extractive summarization, abstractive techniques are more daunting yet more attractive and flexible. Therefore, more and more attention is given to abstractive techniques in different languages. However, to the best of our knowledge, too few works have been dedicated to text summarization in the Persian language, of which almost all are extractive. This is partly due to the lack of proper Persian text datasets available for this task. This is the primary motivation behind the current work: to create an abstractive text summarization framework for the Persian language and compose a new properly formatted dataset for this task.\\
There are different approaches towards abstractive text summarization, especially for the English language, of which many are based on Sequence-to-Sequence (Seq2Seq) structures as text summarization can be viewed as a Seq2Seq task.\\
In \cite{DeepRG} a Seq2Seq encoder-decoder model, in which a deep recurrent generative decoder is used to improve the summarization quality, is presented. The model presented in \cite{AbstractiveTS} is an attentional encoder-decoder Recurrent Neural Network (RNN) used for abstractive text summarization. In \cite{Paulus2018ADR}, a new training method is introduced that combines reinforcement learning with supervised word prediction. An augmented version of a Seq2Seq model is presented in \cite{See2017GetTT}. Similarly, an extended version of encoder-decoder architecture that benefits from an information selection layer for abstractive summarization is presented in \cite{Li2018ImprovingNA}.\\
Many of the works mentioned above benefit from pre-trained language models as these models have started to gain tremendous popularity over the past few years. This is because they simplify each NLP task to a lightweight fine-tuning phase by employing transfer learning benefits. Therefore, an approach to pre-train a Seq2Seq structure for text summarization can be quite promising.\\
BERT \cite{bert}, and T5 \cite{T5} are amongst widely used pre-trained language modeling techniques. BERT uses a Masked Language Model (MLM) and an encoder-decoder stack to perform joint-conditioning on the left and right context. T5, on the other hand, is a unified Seq2Seq framework that employs Text-to-Text format to address NLP text-based problems.\\
A multilingual variation of the T5 model is called mT5 \cite{mt5} that covers 101 different languages and is trained on a Common Crawl-based dataset. Due to its multilingual property, the mT5 model is a suitable option for languages other than English. The BERT model also has a multilingual version. However, there are numerous monolingual variations of this model \cite{arabert,camem} that have shown to outperform the multilingual version on various NLP tasks. For the Persian language, the ParsBERT model \cite{parsbert} has shown state-of-the-art on many Persian NLP tasks such as Named Entity Recognition (NER) and Sentiment Analysis.\\
Although pre-trained language models have been quite successful in terms of Natural Language Understanding (NLU) tasks, they have shown less efficiency regarding Seq2Seq tasks. As a result, in the current paper, we seek to address the mentioned shortcomings for the Persian language regarding text summarization by making the following contributions:

\begin{itemize}
    \item Introducing a novel dataset for the Persian text summarization task. This dataset is publicly available \footnote{http://github.com/hooshvare/pn-summary} for anyone who wishes to use it for any future work.
    \item Investigating two different approaches towards abstractive text summarization for Persian texts. One is to use the ParsBERT model in a Seq2Seq structure as presented in \cite{bert2bert}. The other one is to use the mT5 model. Both models are fine-tuned on the proposed dataset.
\end{itemize}

The rest of this paper is structured as follows. Section \ref{sec:Models} outlines the ParsBERT Seq2Seq encoder-decoder model as well as mT5. In section \ref{sec:configurations}, an overview of the fine-tuning and text generation configurations for both approaches is provided. The composition of the dataset and its statistical features are introduced in section \ref{sec:evaluation}. This section also outlines the metrics used to measure the performance of the models. Section \ref{sec:results} presents the results obtained from fine-tuning the dataset mentioned in earlier models. Finally, section \ref{sec:conclusion} concludes the paper. 

\section{Models}
\label{sec:Models}
In this section, an overview of Sequence-to-Sequence ParsBERT and mT5 architecture is provided.
\subsection{Sequence-to-Sequence ParsBERT}
\label{sec:parsbert}
\begin{figure}[htbp]
\centerline{\includegraphics[width=1.0\linewidth]{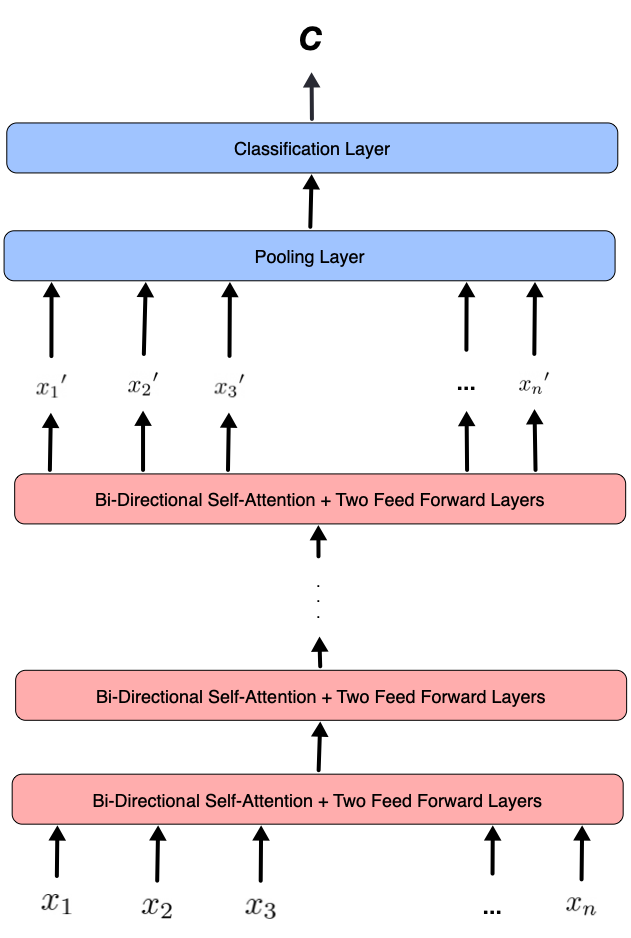}}
\caption{The encoder-only architecture of BERT. Other variations of BERT, such as ParsBERT, have the same architecture.}
\label{fig:BERTARCH}
\end{figure}

ParsBERT \cite{parsbert} is a monolingual version of BERT language model \cite{bert} for the Persian language that adopts the base configuration of the BERT model (i.e. 12 hidden layers, hidden size of 768 with 12 attention heads). BERT is a transformer-based \cite{transformer} language model with an encoder-only architecture that is shown in figure \ref{fig:BERTARCH}. In this architecture the input sequence \(\left \{ x_{1}, x_{2}, ..., x_{n} \right \}\) is mapped to a contextualized encoded sequence \(\left \{ x_{1}^{'}, x_{2}^{'}, ..., x_{n}^{'} \right \}\) by going through a series of bi-directional self-attention blocks with two feed-forward layers in each block. The output sequence can then be mapped to a task-specific output class by adding a classification layer to the last hidden layer.

\begin{figure}[htbp]
\centerline{\includegraphics[width=1.0\linewidth]{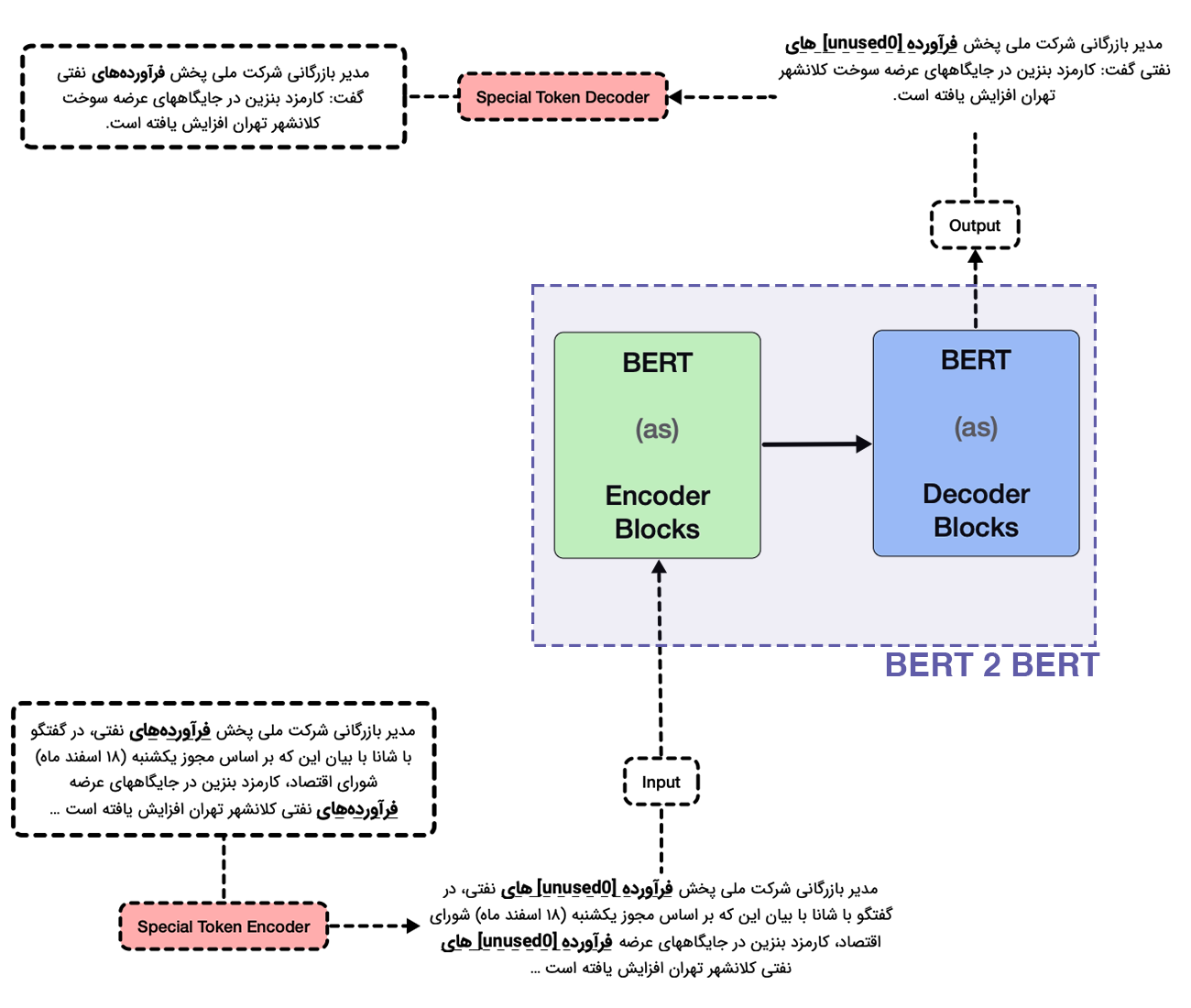}}
\caption{BERT2BERT architecture along with an example Persian text and its summarized version generated by the model.}
\label{fig:bert2bert}
\end{figure}

BERT model achieves state-of-the-art performance on NLU tasks by mapping input sequences to output sequences with a priori known output lengths. However, since the output sequence dimension does not rely on the input, it is impractical to use BERT for text generation (summarization). In other words, any BERT-based model corresponds to the architecture of only the encoder part of transformer-based encoder-decoder models, which are mostly used for text generation.\\
On the other hand, decoder-only models such as GPT-2 \cite{GPT2} can be used as a means of text generation. However, it has been shown that encoder-decoder structures can perform better for such a task \cite{Raffel2020ExploringTL}.\\
As a result, we used ParsBERT to warm-start both encoder and decoder from an encoder-only checkpoint as mentioned in \cite{bert2bert}, to achieve a pre-trained encoder-decoder model (BERT2BERT or B2B) which can be fine-tuned for text summarization using the dataset introduced in section \ref{sec:evaluation}.\\
In this architecture, the encoder layer is the same as the ParsBERT transformer layers. The decoder layers are also the same as that of ParsBERT, with a few changes. First, cross-attention layers are added between self-attention and feed-forward layers in order to condition the decoder on the contextualized encoded sequence (e.g., the output of the ParsBERT model). Second, the bi-directional self-attention layers are changed into uni-directional layers to be compatible with the auto-regressive generation. All in all, while warm-starting the decoder, only the cross-attention layer weights are initialized randomly, and all other weights are ParsBERT's pre-trained weights.\\
figure \ref{fig:bert2bert} illustrates the building blocks of the proposed BERT2BERT model warm-started with the ParsBERT model along with an example text and its summarized version generated by the proposed model.

In this figure, the input text is first fed to a special token encoder that handles half-space character (\textbf{U+200C} Unicode) and removes unwanted tokens. Half-space character is widely used in the Persian language in various situations (e.g. forming plural nouns). In the example text shown in figure \ref{fig:bert2bert}, the word "\RL{fr'Awrdh\nospace hAY}" is actually composed of three tokens: "\RL{fr'Awrdh}" (noun) + [unused0] + "\RL{hAY}" (pluralizing token) where the [unused0] token represents the half-space token that connect the noun to the pluralizing token.\\
After that, the text is fed into the encoder block, the result of the encoder block is fed to the decoder block, which in turn generates the output summary. The half-character tokens are then converted to actual half characters by the particular token decoder block.

\subsection{mT5}
\label{sec:mt5}
mT5 stands for Multilingual Text-to-Text Transfer Transformer (Multilingual T5) and is a multilingual version of the T5 model. T5 is an encoder-decoder Transformer architecture that closely reflects the primary building block of the Original Transformer model \cite{vaswani2017attention} and covers the following objectives:

\begin{itemize}
    \item \textbf{Language Modeling} to predict the next word.
    \item \textbf{De-shuffling} to redefine the original text.
    \item \textbf{Corrupting Spans} to predict masked words.
\end{itemize}

\begin{figure}[htbp]
\centerline{\includegraphics[width=1.0\linewidth]{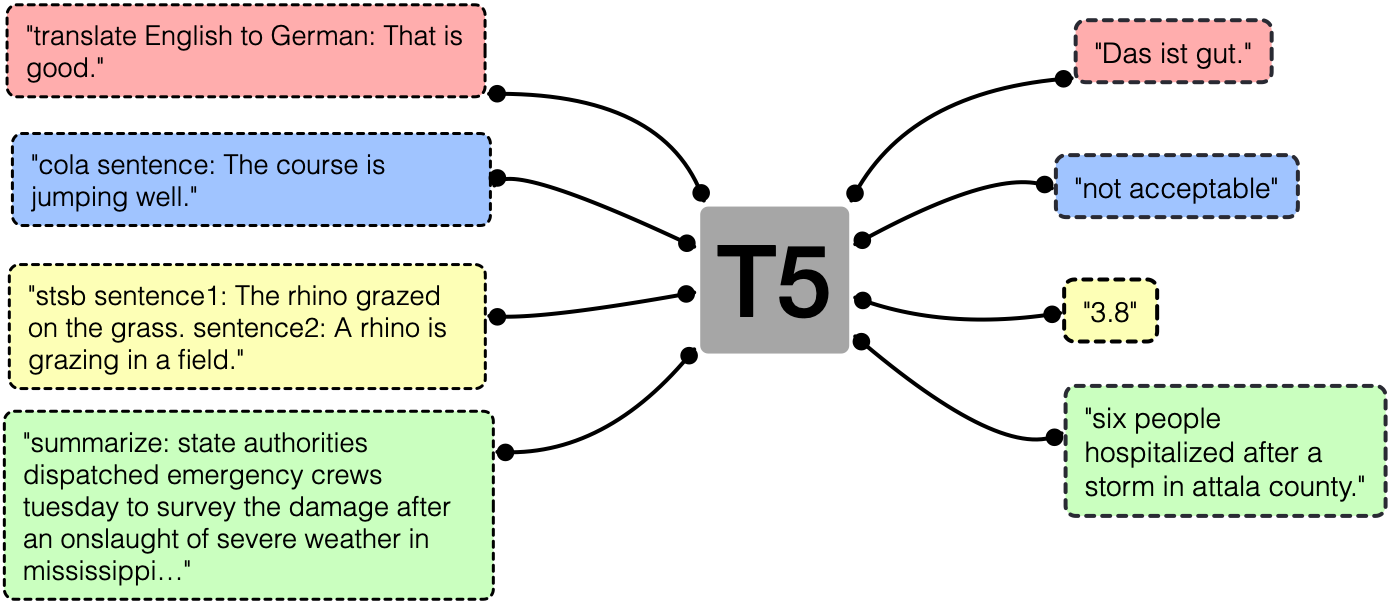}}
\caption{T5 as a unified framework for down-stream NLP tasks. The diagram shows each down-stream task in a text-to-text format, including translation (red), linguistic acceptability (blue), sentence similarity (yellow), and text summarization (green) \cite{Raffel2020ExploringTL}.}
\label{fig:t5-t2t}
\end{figure}

T5 network architecture inherits and transforms the previous unifying frameworks for down-stream NLP tasks into a text-to-text format \cite{Raffel2020ExploringTL}. In other words, the T5 architecture allows for employing the encoder-decoder procedure to aggregate every possible NLP task into one network. Thus, the same hyper-parameters and loss function are used for every task. This is shown in figure \ref{fig:t5-t2t}.

\begin{figure}[htbp]
\centerline{\includegraphics[width=1.0\linewidth]{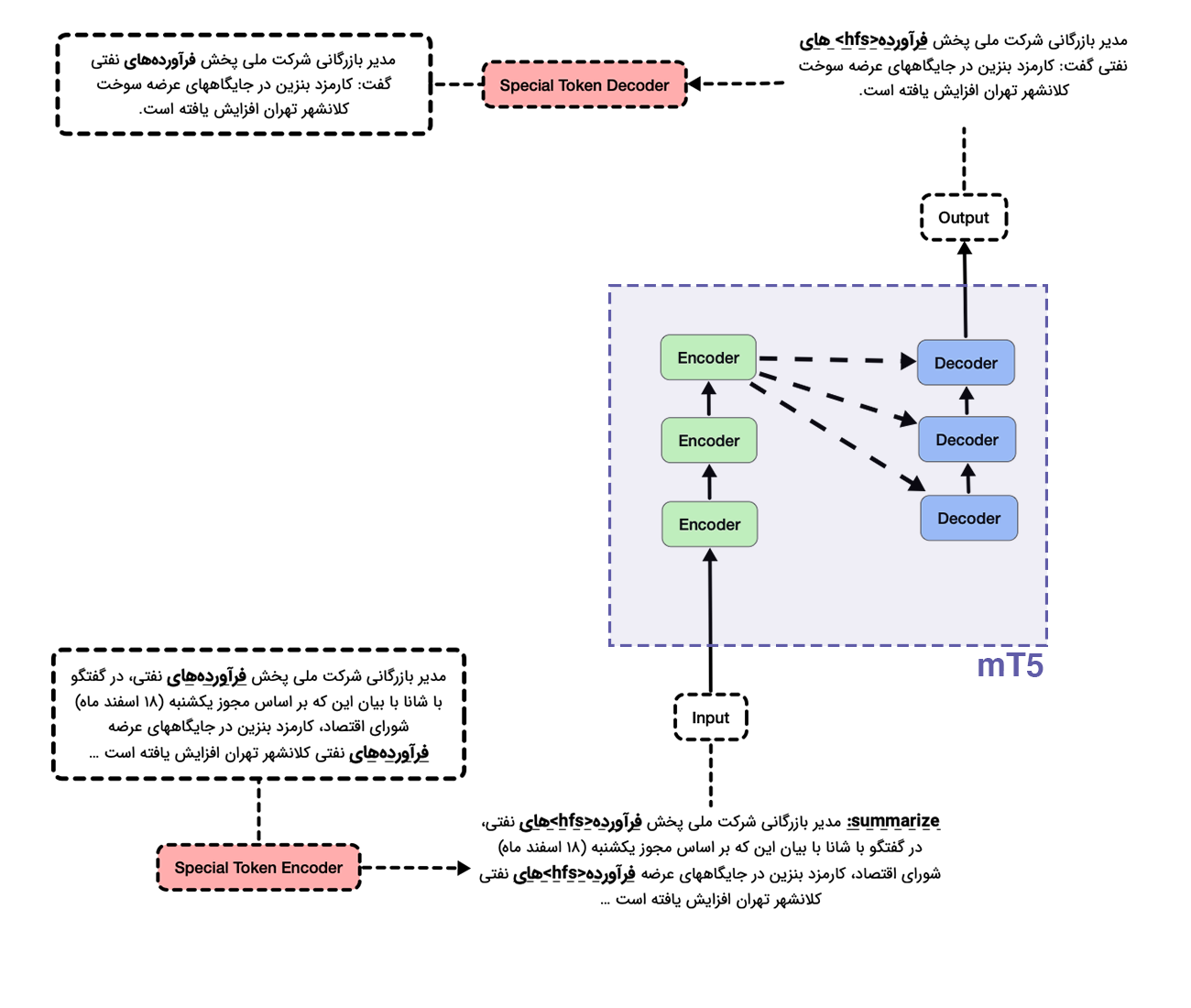}}
\caption{mT5 architecture solution and an example Persian text and its summarized version generated by the model.}
\label{fig:mt5_arch}
\end{figure}

mT5 inherits all capabilities of the T5 model. mT5 was trained on an extended version of the C4 dataset that contains more than 10,000 and web page contents in 101 languages (including Persian) over 71 monthly scrapes to date.\\ 
mT5, compared to other multilingual models like multilingual BERT \cite{bert}, XLM-R \cite{xlm-r}, and multilingual BERT (no support for Persian) \cite{mbart}, reaches state-of-the-art on all the tasks \cite{T5,mt5}, especially on the summarization task.\\
figure \ref{fig:mt5_arch} illustrates the mT5 architecture after fine-tuning, along with an example text. In this schema, the \textbf{<hfs>} token represents the half-space character in Persian and \textbf{"summarize:"} serves as a text-to-text flag for Summarization task.

\section{Configurations}
\label{sec:configurations}
\subsection{Fine-Tuning Configuration}
\label{sec:finetuning}
To Fine-tune both models presented in section \ref{sec:Models} on the pn-summery dataset introduced in section \ref{sec:evaluation}, we have used the Adam optimizer with 1000 warm-up steps, a batch size of 4 and 5 training epochs. The learning rate for Seq2Seq ParsBERT and mT5 are \(5e-5\) and \(1e-4\), respectively.

\subsection{Text Generation Configuration}
\label{sec:textgeneration}
The text generation process refers to the decoding strategy for auto-regressive language generation after the fine-tuned model. In essence, the auto-regressive generation is centered around the assumption that the probability distribution of any word sequence can be decomposed into a product of conditional next word distributions as denoted by equation \eqref{eq:1} where \(W_{0}\) is the initial context word, and \(T\) is the length of the word sequence.

\begin{equation}
P(w_{1:T}|W_{0})=\prod_{t=1}^{T}P(w_{t}|w_{1:t-1},W_{0})
\label{eq:1}
\end{equation}

The objective here is to maximize the sequence probability by choosing the optimal tokens (words). One method is \textit{greedy search} in which the next word selected is simply the word with the highest probability. This method, however, neglects words with high probabilities if they are hidden behind some low probability words.\\
To address this problem, we use \textit{beam search method} that keeps \(n_beams\) number of most likely sequences (i.e., beams) at each time step and eventually chooses the one with the highest overall probability. Beam search generates higher probability sequences as compared to a greedy search.\\
One drawback is that beam search tends to generate sequences with some words repeated. To overcome this issue, we utilize n-grams penalties \cite{Paulus2018ADR, Klein2017OpenNMTOT}. This way, if a next word causes the generation of an already seen n-grams, the probability of that word will be set to \(0\) manually, thus preventing that n-gram from being repeated. Another parameter used in beam search is early stopping, which can be either active or inactive. If active, text generation is stopped when all beam hypotheses reach the EOS token. The number of beams, the n-grams penalty sizes, the length penalty and early stopping values used for BERT2BERT and mT5 models in the current work are presented in table \ref{table:beamsearch}.\\

\begin{table}[htbp]
\centering
\caption{Beam search configuration for BERT2BERT and mT5 models for auto-regressive text summarization after fine-tuning.}
\label{table:beamsearch}
\begin{tabular}{@{}ccc@{}}
                 \midrule
                & BERT2BERT & mT5 \\ \midrule
    \# Beams     & 3 & 4 \\\midrule
    Repetitive N-gram Size \cite{Paulus2018ADR,Klein2017OpenNMTOT}     & 2 & 3 \\\midrule
    Length Penalty\cite{Paulus2018ADR,Klein2017OpenNMTOT}     & 2.0 & 1.0 \\\midrule
    Early Stoping Status & ACTIVE & ACTIVE \\\bottomrule
\end{tabular}%
\end{table}

\section{Evaluation}
\label{sec:evaluation}
\begin{figure}[htbp]
\centerline{\includegraphics[width=1.0\linewidth]{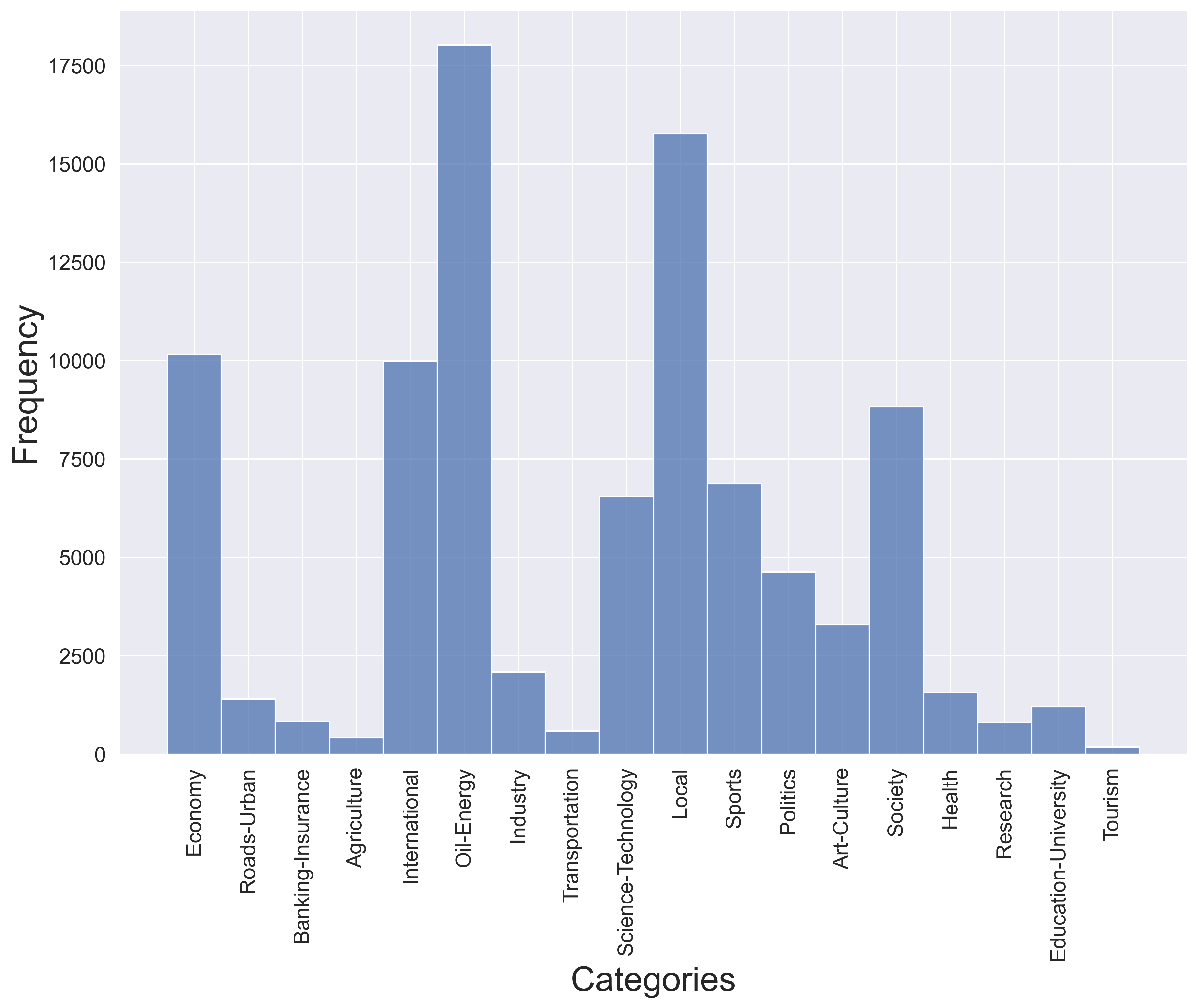}}
\caption{The frequency of article categories in the proposed dataset.}
\label{fig:datasetcategory}
\end{figure}

\begin{figure}[htbp]
\centerline{\includegraphics[width=0.8\linewidth]{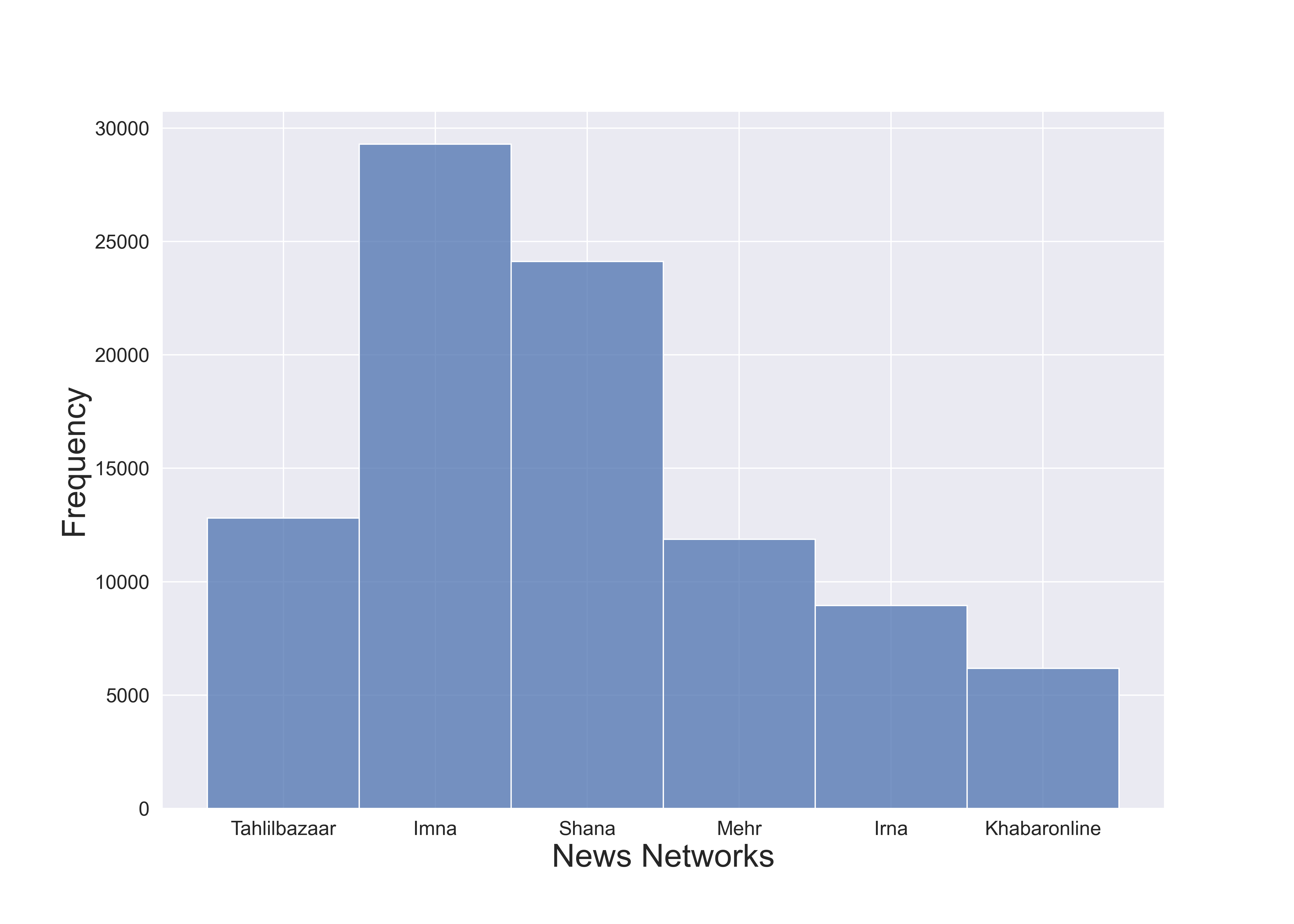}}
\caption{The number of articles extracted from each of the news agency website.}
\label{fig:datasetagency}
\end{figure}

For evaluating the performance of the two architectures introduced in this paper, we composed a new dataset by crawling numerous articles along with their summaries from 6 different news agency websites, hereafter denoted as pn-summary. Both models are fine-tuned on this dataset. Therefore, this is the first time this dataset is being proposed to be used as a benchmark for Persian abstractive summarization. This dataset includes a total of 93,207 documents and covers a range of categories from economy to tourism. The frequency distribution of the article categories and the number of articles from each news agency can be seen from figures \ref{fig:datasetcategory} and \ref{fig:datasetagency}, respectively.

\begin{figure}[htbp]
\centerline{\includegraphics[width=1.0\linewidth]{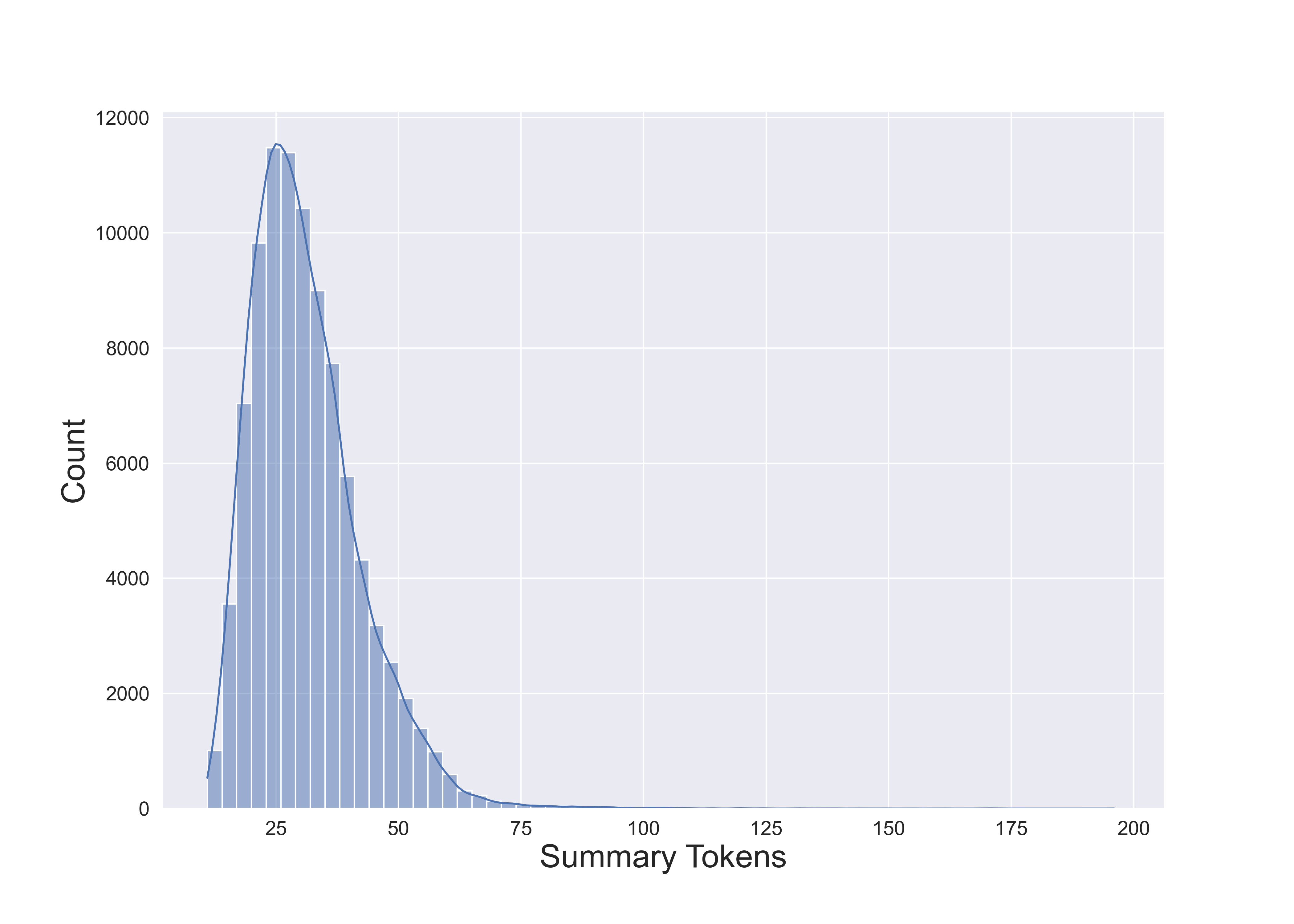}}
\caption{Token length distribution of articles' summaries.}
\label{fig:datalength}
\end{figure}

It should be noted that the number of tokens in article summaries is varying. This can be viewed in figure \ref{fig:datalength}. As shown from this figure, most of the articles' summaries have a length of around 30 tokens.

To determine the performance of the models, we use Recall-Oriented Understudy for Gisting Evaluation (ROUGE) metric package \cite{rouge}. This package is widely used for automatic summarization and machine translation evaluation. The metrics included in this package compare an automated summary against a reference summary for each document. There are five different metrics included in this package. We calculate the F-1 score for three of these metrics to show the overall performance of both models on the proposed dataset:

\begin{itemize}
\item \textbf{ROUGE-1 (unigram) scoring} which computes the overlap of uni-grams between the generated and the reference summaries.
\item \textbf{ROUGE-2 (bigram) scoring} which computes the overlap of bigrams between the generated and the reference summaries.
\item \textbf{ROUGE-L scoring} in which the scores are calculated at sentence-level. In this metric new lines are ignored, and Longest Common Subsequence (LCS) is computed between two text pieces.
\end{itemize}

\section{Results and Discussion}
\label{sec:results}
This section presents the results obtained from fine-tuned mT5 and ParsBERT-based BERT2BERT structure on the proposed pn-summary dataset. The $F_{1}$ scores on three different ROUGE metrics discussed in section \ref{sec:evaluation} are reported in table \ref{table:result_rouge}. It can be seen that the ParsBERT B2B structure achieves higher scores as compared to the mT5 model. This could be due to the fact that encoder-decoder weights (i.e., ParsBERT weights) in this architecture are concretely tuned on a massive Persian corpus, making it a fitter architecture for Persian-only tasks.

\begin{table}[htbp]
\centering
\caption{Depicts ROUGE-F1 scores on the test set. The objective of models and baselines is abstractive. The two models are fine-tuned on the Persian news summarization dataset (pn-summary).}
\label{table:result_rouge}
\begin{tabular}{@{}cccc@{}}
    \toprule
                & \multicolumn{3}{c}{ROUGE} \\ \midrule
    Model       & R-1 & R-2 & R-L \\ \midrule
    mT5         & 42.25 & 24.36 & 35.94  \\ \midrule
    BERT2BERT   & \textbf{44.01} & \textbf{25.07} & \textbf{37.76} \\ \bottomrule
\end{tabular}%
\end{table}

Since no other pre-trained abstractive summarization methods have been proposed for Persian language and since this is the first time the pn-summary dataset is being introduced and released, it is impossible to compare the results of the present work with any other baseline. As a result, the outcomes presented in this work can serve as a baseline for any future abstractive methods for the Persian language that seeks to train their model on the proposed pn-summary dataset presented and released with the current work.\\
To further illustrate these two models' performance, we have included two examples from the dataset in table \ref{table:result_generation_mt5}. The main text, the actual summary, and the summaries generated by the mT5 and BERT2BERT models are shown in this table. Based on this table, the summary given by the BERT2BERT model in both examples is relatively closer to the actual summary in terms of both meaning and lexical choices.

\begin{table}[htbp]
\centering
\caption{Examples of highly abstractive reference summaries from Persian News Network using mT5 and BERT2BERT (B2B) models. Each example consists of the trim article, the true summary, and the generated summaries by both models.}
\label{table:result_generation_mt5}
\resizebox{\linewidth}{!}{%
\begin{tabular}{>\justify m{0.80\linewidth}c}
    \toprule
    \centering
    Example & \# \\\midrule
    
    \begin{RLtext}
    \footnotesize{\textbf{mtn _hbr:} bh gzAr^s _hbrngAr bAzAr, mfId .glAmY .sb.h pn^g^snbh dr n^sst stAd Aqt.sAd mqAwmtY dr sAln AstAndArY mAzndrAn bA A^sArh bh r^sd 36 dr.sdY w.swl dr'AmdhAY AstAn, mIzAn dr'AmdhAY m.swb sAl ^gArY rA 10 hzAr mIlIArd rIAl A`lAm .krd  [...]}
    \end{RLtext}
    \begin{RLtext}
    \footnotesize{\textbf{_hlA.sh A.slY:} sArY - ryIs sAzmAn mdIrIt w brnAmh\nospace rIzY mAzndrAn mIzAn dr'AmdhAY m.swb `mwmY sAl ^gArY dr AstAn rA 10 hzAr mIlIArd rIAl A`lAm .krd.}
    \end{RLtext} & (1) \\ \hdashline
    \begin{RLtext}
    \footnotesize{ryIs sAzmAn mdIrIt w brnAmh\nospace rIzY mAzndrAn gft: bh r.gm w^gwd AIn ^srAI.t 86 dr.sd shm dr'Amd AstAn Az dr'AmdhAY mAlIAtY m.hq.q ^sdh Ast.}
    \end{RLtext} & \textbf{mT5} \\ \hdashline
    \begin{RLtext}
    \footnotesize{sArY - rIIs sAzmAn mdIrIt w brnAmh\nospace IzY mAzndrAn mIzAn drAmdhAY m.swb sAl ^gArY AstAn rA 10 hzAr mIlIArd rIAl _d.kr .krd w gft: AstAn ptAnsIl w ^srAI.t mnAsbY dr .hwzh\nospace hAY m_htlf dArd.}
    \end{RLtext} & \textbf{B2B} \\\midrule
    
    \begin{RLtext}
    \footnotesize{\textbf{mtn _hbr:} bh gzAr^s AIrnA; `.sr ^snbh Az hfth yAzdhm lIg brtr bA^sgAh\nospace hAY AnglIs, mn^cstrsItY dr wrz^sgAh At.hAd Az fwlAm p_dIrAIY .krd w .hrIf AnthAY ^gdwlY _hwd rA bA .hsAb dw br .sfr Az pI^s\nospace rwY brdA^st tA bh pn^gmIn brd f.sl brsd. sItY .kh bAzY rA thA^gmY 'A.gAz .krd, _hIlY zwd w dr dqIqh pn^gm bA .drbh r.hIm AstrlIng bh gl dst yAft (...)}
    \end{RLtext} 
    \begin{RLtext}
    \footnotesize{\textbf{_hlA.sh A.slY:} tIm fwtbAl mn^cstrsItY dr dIdAr _hAngY mqAbl fwlAm bh brtrY dw br .sfr dst yAft.}
    \end{RLtext} & (1) \\\hdashline
    \begin{RLtext}
    \footnotesize{tIm mn^cstrsItY dr wrz^sgAh At.hAd Az fwlAm p_dIrAIY .krd w .hrIf AnthAY ^gdwlY _hwd rA bA .hsAb dw br .sfr Az pI^s rwY brdA^st.}
    \end{RLtext} & \textbf{mT5} \\\hdashline
    \begin{RLtext}
    \footnotesize{tIm fwtbAl mn^cstrsItY dr dIdAr _hAr^g Az _hAnh mqAbl mIhmAn _hwd bh brtrY dw br .sfr dst yAft tA bh pn^gmIn brd f.sl brsd.}
    \end{RLtext}  & \textbf{B2B} \\\midrule
\end{tabular}%
}
\end{table}

\section{Conclusion}
\label{sec:conclusion}
Limited work has been dedicated to text summarization for the Persian language, of which none are abstractive based on pre-trained models. In this paper, we presented two pre-trained methods and designed to address text summarization in Persian with an abstract approach: one is based on a multilingual T5 model, and the other is a BERT2BERT warm-started from the ParsBERT language model. We have also composed and released a new dataset called pn-summary for text summarization since there is an apparent lack of such datasets for the Persian language. The results of fine-tuning the proposed methods on the mentioned dataset are promising. Due to a lack of works in this area, our work could not be compared to any earlier work and can now serve as a baseline for any future works in this field.

\bibliographystyle{unsrt}
\bibliography{references}
\vspace{12pt}

\end{document}